# A Neural Network Training Method Based on Distributed PID Control

Jiang Kun

Chongqing university

atan_j@qq.com

**Abstract**: In the previous article, we introduced a neural network framework based on symmetric differential equations. This novel framework exhibits complete symmetry, endowing it with perfect mathematical properties. While we have examined some of the system's mathematical characteristics, a detailed discussion of the network training methodology has not yet been presented. Drawing on the principles of the traditional backpropagation algorithm, this study proposes an alternative training approach that utilizes differential equation signal propagation instead of chain rule derivation. This approach not only preserves the effectiveness of training but also offers enhanced biological interpretability. The foundation of this methodology lies in the system's reversibility, which stems from its inherent symmetry—a key aspect of our research. However, this method alone is insufficient for effective neural network training. To address this, we further introduce a distributed Proportional-Integral-Derivative (PID) control approach, emphasizing its implementation within a closed system. By incorporating this method, we achieved both faster training speeds and improved accuracy. This approach not only offers novel insights into neural network training but also extends the scope of research into control methodologies. To validate its effectiveness, we apply this method to the MNIST dataset, demonstrating its practical utility.

## 1. Introduction

In previous research, we proposed a novel neural network architecture that integrates the symmetrical Five Elements (Wuxing) Theory with the asymmetrical predator-prey equation, resulting in a set of fully symmetrical differential equations[1]. This complete symmetry endows the system with favorable mathematical properties, such as the ability to specify fixed points and reversible propagation directions, both of which preserve the system's causality. Specifying fixed points enables the system to operate in a desired state, while reversible propagation enhances the network's training capabilities. Although we briefly outlined the training method for this neural network in the previous article, we did not provide a detailed theoretical foundation or explanation. In this article, we offer a comprehensive introduction to the training methodology, including both the underlying theory and a distributed Proportional-Integral-Derivative (PID) training approach, aimed at achieving higher accuracy.

The training of neural networks is a highly complex problem that involves various aspects, including strategy formulation, method implementation, structural design, and parameter adjustment. This complexity has led to the development of a wide range of neural network training methods. Among these, the traditional backpropagation algorithm remains the dominant approach. While backpropagation has achieved significant success, its biological plausibility has been widely questioned[2]. As a result, numerous alternative algorithms have been proposed to replace backpropagation, although they have yet to gain widespread adoption.

To develop an effective alternative to the backpropagation algorithm, it is essential to first understand the fundamental strengths of backpropagation. In our view, the key advantage of the backpropagation method lies in its point-to-point, precise adjustment capability, which contributes to both training efficiency and robustness. By using backpropagation, we can quantitatively assess the impact of each parameter on the final outcome, enabling targeted adjustments—a feature that many alternative algorithms fail to provide. Therefore, despite criticisms regarding its biological plausibility, the backpropagation algorithm remains indispensable and cannot easily be replaced by other methods.

In our effort to propose an efficient alternative, we have approached the problem from both mathematical and biological perspectives.

From a mathematical perspective, we propose using signal propagation in differential equations as a replacement for chain rule derivation, a method we refer to as differential equation propagation. This approach maintains a one-to-one correspondence, ensuring high efficiency. The key prerequisite for this method is that the system must be reversible. As established in our earlier design, the system exhibits complete symmetry, which makes reversibility straightforward to achieve. Reversibility implies that we can trace the causal relationships within the system in reverse, analogous to the point-to-point mechanism in the backpropagation algorithm.

From a biological standpoint, we introduce the concept of instinctive design. In this framework, each neuron operates autonomously and retains full functionality, meaning that individual neurons can independently execute signal propagation and feedback training without the need for global information. Neurons interact with the external environment through synapses, without requiring any specialized structural design. This approach provides both strong biological interpretability and efficient training capabilities. Within this training framework, each neuron adjusts system parameters by comparing forward and reverse signals passing through it, thereby facilitating neural network training without the need for global coordination.

However, adopting this strategy alone is insufficient. The system composed of differential equations contains three distinct sets of parameters. If the same adjustment strategy were applied to all three parameter sets during training, it would be ineffective in enhancing neural network performance. To address this issue, we introduce a distributed Proportional-Integral-Derivative (PID) control method. PID control holds a dominant position in traditional control systems, whether in academic teachings or industrial applications, and remains a critical tool[3]. However, our approach deviates from the conventional PID method. In alignment with the principle of instinctive design, we focus on implementing the PID control logic within a single closed system. Building on the fundamental concepts of PID control, we have developed a novel control method that not only adheres to the instincts of design but also effectively incorporates PID functionality.

To validate our approach, we conducted experiments on the MNIST dataset. The results demonstrate that the distributed PID control method effectively enhances the system's accuracy and training speed.

The remainder of this article is organized as follows: Chapter 2 provides a brief introduction to the Wuxing neural network, offering a foundational understanding of its core concepts. Chapter 3 details the training methodology for the Wuxing neural network, emphasizing the use of differential equation signal propagation as a replacement for chain rule derivation. Chapter 4 explores PID control theory and its application to neural network training, with a particular focus on implementing distributed PID methods in closed systems. Chapter 5 concludes with a summary of our work and prospects for future research.

## 2. Wuxing neural network

In this section, we will briefly introduce the Wuxing neural network structure, fixed point calculation, and signal propagation method. For more details, please refer to our previous article[1].

## 2.1 Wuxing neural network differential equations

Traditional neural networks can generally be divided into two categories: those based on mathematical principles, such as multilayer perceptron (MLPs) and Hopfield networks[4]. And those inspired by biological systems, such as chaotic neural networks and cellular neural networks[5, 6]. Mathematically driven neural networks have given rise to models like convolutional neural networks (CNNs) and recurrent neural networks (RNNs), which serve as the foundation for many large-scale, widely deployed models today. In contrast, while biologically inspired neural networks offer strong biological interpretability, they lack scalability for practical deployment. This limitation stems from their reliance on differential equations, for which suitable mathematical formulations to accurately describe neural activity are still lacking.

In the Wuxing neural network, neurons are modeled as systems composed of a series of symmetrical differential equations. The primary objective of introducing this structure is to address the challenge of manipulating differential equations effectively. Traditional biological neural networks often employ differential equations derived from experimental observations. However, this approach can introduce detailed inconsistencies that undermine the equations' mathematical properties. To address this issue, our research adopts symmetrical logic based on the Five Elements (Wuxing) theory as its foundation. By incorporating elements of the predator-prey equation, we develop a set of fully symmetrical differential equations to model neural activities, replacing conventional neurons with systems governed by these equations.

Fig. 1 From Wuxing logic to symmetric differential equations

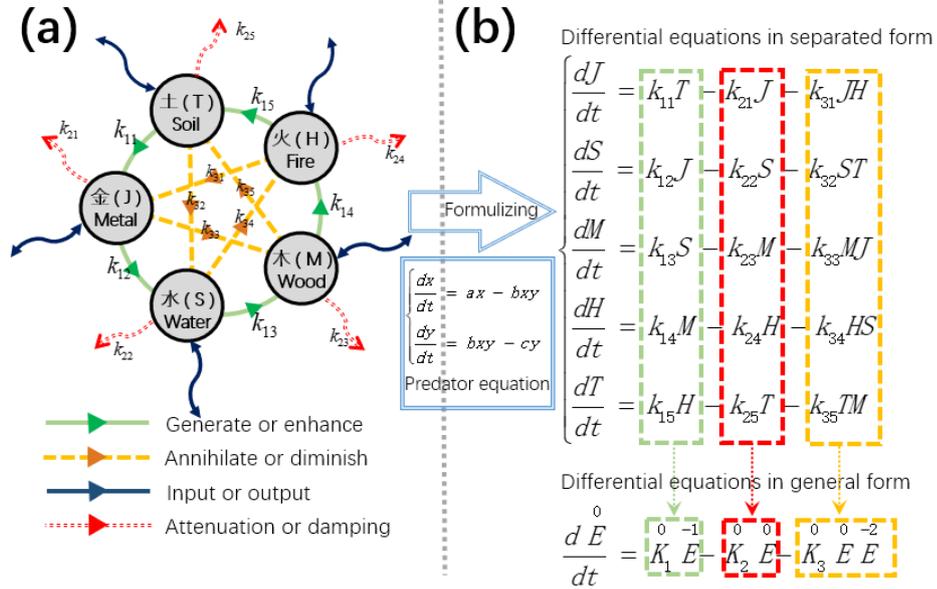

Fig. 1. a Traditional Wuxing logic posits that the world is composed of five distinct elements that interact through generative and inhibitory relationships, forming the logical framework of the universe. The logical structure depicted in the figure deviates from the traditional model by incorporating self-attenuation terms and creating interfaces for external input and output signals. In this system, there are five nodes, each capable of serving as an input or output. However, to prevent signal interference, a node can either receive input or generate output at any given time, but not both simultaneously.

Fig. 1. b By combining the Wuxing logic with the predator-prey equation, we can derive a set of differential equations. The symmetry of the system is carefully preserved throughout this transformation process. As a result, both the traditional Five Elements logic and the predator-prey equation are modified, ultimately leading to a set of fully symmetrical equations. For clarity, these equations are presented in a generalized format, with the numbers above the elements and parameters indicating the offset of each loop.

Figure 1a illustrates the logical structure underpinning the differential equations, while Figure 1b presents their mathematical formulation. According to the Wuxing theory, the world is composed of five distinct elements that interact through generative and inhibitory relationships, forming a completely symmetrical logical framework. Building on this conceptual basis and incorporating the predator-prey equation, we derived the equations represented in Figure 1b.

In order to better describe similar equations, we use the following general formula to describe the original system of equations:

$$\frac{dE}{dt} = K_1 E - K_2 E - K_3 EE \tag{2.1}$$

In order to represent elements in different orders, we define $E = \{J, S, M, H, T\}$, $K_1 = \{k_{11}, k_{12}, k_{13}, k_{14}, k_{15}\}$, $K_2 = \{k_{21}, k_{22}, k_{23}, k_{24}, k_{25}\}$, $K_3 = \{k_{31}, k_{32}, k_{33}, k_{34}, k_{35}\}$, so the Wuxing differential equation above can be expressed as:

$$\frac{d \overset{0}{E}}{dt} = \overset{0}{K_1} \overset{-1}{E} - \overset{0}{K_2} \overset{0}{E} - \overset{0}{K_3} \overset{0}{E} \overset{-2}{E} \tag{2.2}$$

The number above the variable indicates the offset of the loop: $\overset{-1}{E} = \{T, J, S, M, H\}$.

## 2.2 Fixed points of differential equations

In chaos theory, the fixed point of a differential equation is a critical property, as it determines the equilibrium state of the system. In Equation 2.2, three distinct sets of parameters are involved. To analytically determine the fixed point of the equation, it is necessary to simplify the parameters. We assume that the parameters within each of the three sets—denoted as $K_1$, $K_2$ and $K_3$—are equal, allowing us to analytically derive the fixed point B0 of the equation.

$$B_0 = \frac{K_1 - K_2}{K_3} \tag{2.3}$$

Equation 2.3 plays a central role in adjusting the differential equation. Through this equation, we can modify the fixed point of the system. Even when the parameters in $K_1$, $K_2$ and $K_3$ no longer satisfy the condition of equality, the adjustment method remains effective. Furthermore, Equation 2.3 establishes the relationship between the system's parameters and the fixed point. For instance, increasing $K_1$ and decreasing $K_3$ produces the same effect on the fixed point.

## 2.3 Signal propagation and network structure

In our study, we regard the state where the differential equation is at a fixed point as the zero state of the system. Therefore, we can get the input equation of the system:

$$\frac{d \overset{0}{E(t)}}{dt} = \overset{0}{K_1} \overset{-1}{E(t)} - \overset{0}{K_2} \overset{0}{E(t)} - \overset{0}{K_3} \overset{0}{E(t)} \overset{-2}{E(t)} + Input(t) \tag{2.4}$$

In 2.4, $Input(t)$ is the input signal. Similarly, we can also get the output equation of the system:

$$D(t) = E(t) - B_0 \tag{2.5}$$

In 2.5, $D(t)$ is the output signal, which reflects the magnitude of the system's deviation from the fixed point. When we have both input and output signals, we can further define the network links of the system:

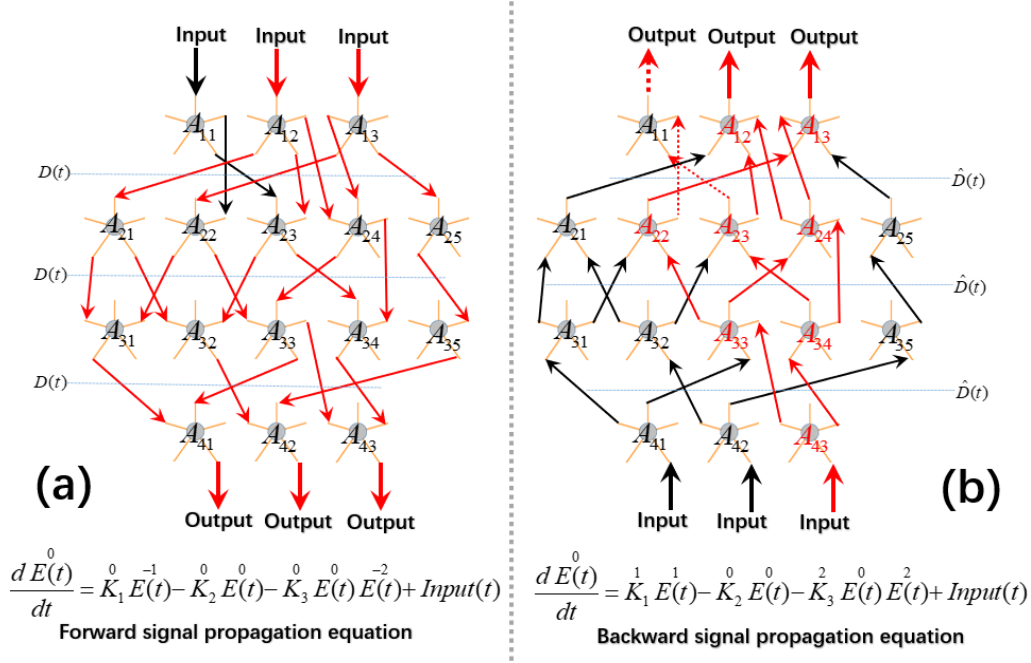

Fig.2 Forward Propagation and Backward Propagation

$$\frac{dE^0(t)}{dt} = K_1^0 E^{-1}(t) - K_2^0 E^0(t) - K_3^0 E^0(t) E^{-2}(t) + Input(t)$$
**Forward signal propagation equation**

$$\frac{dE^0(t)}{dt} = K_1^1 E^1(t) - K_2^0 E^0(t) - K_3^2 E^0(t) E^2(t) + Input(t)$$
**Backward signal propagation equation**

Fig. 2. a Forward propagation signal, red represents signal propagation, black represents no signal propagation. In the figure, $A_{11}$ has no signal input, while $A_{12}$ and $A_{13}$ have signal input.

Fig. 2. b Backward propagation signal, red indicates signal propagation, black indicates no signal propagation. We assume that the outputs of $A_{41}$ and $A_{42}$ meet the set conditions in the previous forward propagation, so only $A_{43}$ has a feedback error signal. Although there is a feedback input signal in A11, there is no input signal in the forward input, so the parameters in $A_{11}$ will not be updated. Only neurons with both forward and reverse inputs (marked in red) will update parameters.

In Figure 2, we have built a 4-layer network with three inputs and three outputs. The connections between neurons in the figure are uniformly random. The signal propagated in the forward network is marked as $D(t)$, and the signal propagated in the reverse network is marked as $\hat{D}(t)$. Similar to the previous example, $\hat{D}(t)$ is determined by the back propagation element value $\hat{E}(t)$ and the back-propagation fixed point $\hat{B}_0$.

$$\hat{D}(t) = \hat{E}(t) - \hat{B}_0 \qquad (2.6)$$

## 3. Training Wuxing neural network

In this section, we will introduce how to use differential equations for signal propagation and achieve point-to-point parameter adjustment.

### 3.1 Training theory

In neural network training, the traditional backpropagation algorithm has been highly successful. Despite ongoing doubts regarding its biological plausibility, no alternative methods have yet been able to match the accuracy achieved by backpropagation. The key advantage of the backpropagation algorithm lies in its point-to-point training approach, which establishes a one-to-one relationship between each parameter and the output. Therefore, when considering the development of a new training method, it is crucial to preserve this point-to-point relationship. From a mathematical perspective, this relationship embodies a form of symmetry, but maintaining this symmetry in practice is not straightforward.

Due to various practical constraints, the equations we derive often lose this symmetry. To address this, we begin with the principle of symmetry and construct symmetric differential equations, which ensure that the system remains symmetric and, consequently, reversible. This reversibility allows us to trace the cause-and-effect relationships in the system by reversing the signal propagation, enabling us to study the influence of parameters through signal flow. In essence, we replace the reverse differentiation process with the use of differential equations. For example, consider the differential system dX=aY. To study the effect of the parameter a on the outcome, we can establish an inverse differential system dY=aX, and compare the results between the two systems to analyze the impact of a on the output.

**3.2  Training method**

According to our definition, the system's signals can propagate in both directions. Based on the system's topological structure, we define the inverse equation of the system:

$$\frac{d \overset{0}{E}(t)}{dt} = \overset{1}{K_1} \overset{1}{E}(t) - \overset{0}{K_2} \overset{0}{E}(t) - \overset{2}{K_3} \overset{0}{E}(t) \overset{2}{E}(t) + Input(t) \tag{3.1}$$

Equation 3.1 and Equation 2.2 are symmetrical of each other. In 3.1, the orders of $K_1$, $K_2$ and $K_3$ parameters are also adjusted. This is because $K_1$, $K_2$ and $K_3$ parameters are actually parameters located between two elements (Figure 1.a). Therefore, when the signal propagation direction changes, the order of parameters must also change.

Equation 3.1 and Equation 2.2 seem to be very different, but they have one thing invariant, that is, the connection between the two elements. This is the core method of training in this paper, which leads to a strong correlation between the forward propagation signal and the reverse propagation signal. This correlation is the core of adjusting the system parameters. In the traditional back-propagation training method, people use the chain rule to find the partial derivative of the parameter to the error to adjust the system, and this derivation process is to find the correlation between the parameter and the signal. Therefore, if we do not use the back-propagation method of derivation, then we must first explain how our method preserves the correlation of the system, and secondly how to apply this correlation to the system and achieve the desired results by changing the parameters.

More generally speaking, finding partial derivatives is also looking for a reversible causal relationship. If the forward input signal and the reverse input signal are located at the two ends of the system, then according to the propagation of the two signals, the signal connection can be established at different nodes. Depending on the way the coefficients are adjusted, different types of connections can be established within the system.

Through this reversible causal relationship, we can achieve many functions. The following is an example. In Figure 2, we have built a 4-layer network with three inputs and three outputs. The signal transmitted in the forward network is marked as $D(t)$, and the signal transmitted in the reverse network is marked as $\hat{D}(t)$. Therefore, we can define an output variable $Leb$ within time $T$ and take the largest $Leb$ component as the output result.

$$Leb = \frac{1}{T} \int_0^T D(t) dt \tag{3.2}$$

Assuming that the $P_{th}$ component of $Leb$ should be the largest, according to our previous research, the input signal for backpropagation can be defined as follows[1]:

For the $P_{th}$ component, the adjustment error is:

$$Error_p = \begin{cases} targtet1 - Leb_p & (if\ Leb_p < target1) \\ 0 & (if\ Leb_p > target1) \end{cases} \tag{3.3}$$

For other component, the adjustment error is:

$$Error_{other} = \begin{cases} targtet2 - Leb_{other} & (if\ Leb_{other} > target2) \\ 0 & (if\ Leb_{other} < target2) \end{cases} \quad (3.4)$$

In these equations, $target1$ and $target2$ represent two predefined target values, where $target1$ is the larger value and $target2$ is the smaller one. This method will make the value of the $P_{th}$ component larger after training, while the others will be smaller, allowing the system to achieve a higher accuracy rate.

Similar to before, $\hat{D}(t)$ is determined by the back-propagated element value $\hat{E}(t)$ and the back-propagated fixed point $\hat{B}_0$.

$$\hat{D}(t) = \hat{E}(t) - \hat{B}_0 \quad (3.5)$$

Assuming that in the forward network of Figure 2, only $A_{12}$ and $A_{13}$ have input signals, then the signal propagation diagram is shown in Figure 2a, where only the paths marked in red have signal propagation. When the signal reaches the output end, it can be compared with the set target. If the comparison is successful, no error signal is returned. If the comparison is unsuccessful, the corresponding error signal is returned. In Figure 2a, we assume that only the result of $A_{43}$ does not meet the requirements, so in Figure 2b, only $A_{43}$ has an input error signal. Based on the forward and reverse signals propagated within time $T$, we can define a correlation variable $G_1$.

$$G_1 = \int_0^T D(t)dt \bullet \int_0^T \hat{D}(t)dt \quad (3.6)$$

Since $G_1$ may exceed a certain limit, we use the inverse tangent function (other similar functions are also possible) to limit it and get $G_2$

$$G_2 = atan(G_1 * kt) / kt \quad (3.7)$$

$kt$ is the adjustment parameter, $G_2$ is the adjusted correlation value. The parameter can be adjusted based on $G_2$.

$$K_{3\_new} = K_{3\_old} \bullet exp(-G_2) \quad (3.8)$$

## 4. Distributed PID Control

In this section, we will discuss how to implement distributed PID control in closed systems to address parameter redundancy issues encountered in neural network training.

### 4.1 Redundant parameter adjustment

In the Wuxing neural network, there are three sets of parameters $K_1$, $K_2$ and $K_3$; in previous studies, we only gave the method to adjust $K_3$ because the method is not reusable. The following is an example trained on the MNIST dataset. The model has 784 inputs, 10 outputs, and a total of 6 layers. The number of neurons in each layer is: {784, 839, 283, 96, 32, 10}, of which the first and last layers are fully connected, and all interfaces have inputs or outputs. The initial parameters of the model are $K_1$={1, 1, 1, 1, 1}; $K_2$={0.5, 0.5, 0.5, 0.5, 0.5}; and $K_3$={0.5, 0.5, 0.5, 0.5, 0.5}. We use the training method in Chapter 3. One is to adjust only $K_3$, and the other is to adjust $K_1$ and $K_3$ at the same time. The method of adjusting K1 can refer to formula 3.5. Combined with the relationship in 2.1, we can get:

$$K_{1\_new} = K_{1\_old} \bullet exp(G_2) \quad (4.1)$$

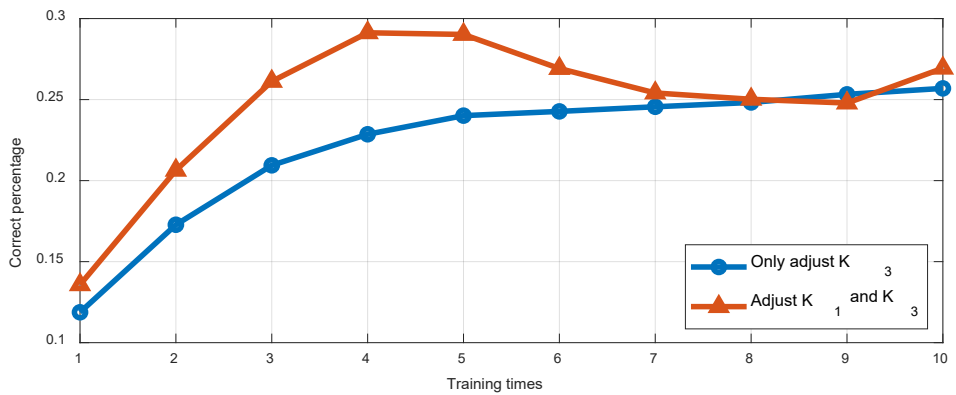

Fig. 3   Accuracy curve based on the forward and backward signal comparison method

Fig. 3 In Case 1, we only adjust K3, and in Case 2, we adjust both K1 and K3. In Case 1, the accuracy rate continues to improve, although the speed is slower than that in Case 2, but the stability is better. In Case 2, although the accuracy rate improves faster, after the fifth training, the accuracy rate begins to decline and fluctuates.

In Figure 3, Case 1 only adjusts $K_3$, and Case 2 adjusts $K_1$ and $K_3$ at the same time. The accuracy of Case 1 improves more slowly, but is also more stable, while the improvement of Case 2 is rapid, but has greater volatility. In this case, simply relying on the same method to adjust two different parameters will not produce the ideal results imagined. This is because the system is strongly coupled. According to Equation 2.3, adjusting two parameters at the same time actually has a similar effect to adjusting only one parameter, and cannot achieve the ideal purpose. At this time, we need to look at the adjustment problem from the perspective of the system.

**4.2  Typical PID control method**

In the automatic control system, PID control is the absolute leader. PID control adjusts the input by feeding back the error signal and finally achieves the ideal control result.

Figure 4 illustrates a typical PID control process. Assuming the input signal represents the target value we set, the corresponding error signal is derived by comparing the output signal with the input signal. The error signal is then processed through proportional, integral, and derivative control, before being re-input into the system. This results in an adjusted signal. After several iterations of parameter adjustments, the desired outcome is typically achieved. However, an important question arises: why is a feedback loop necessary in this process?

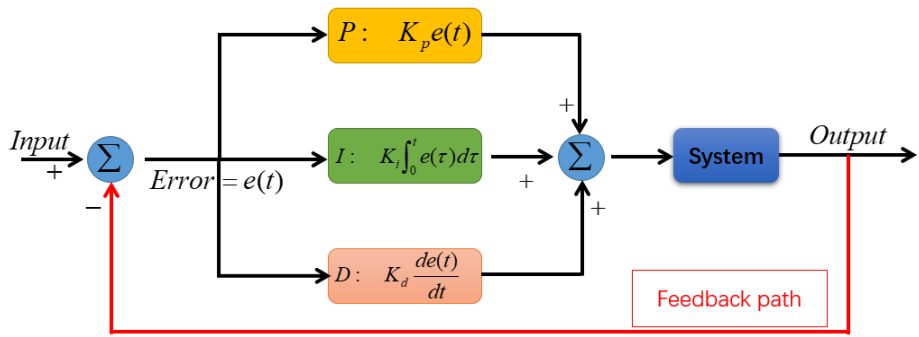

Fig. 4 Typical PID control system

Fig. 4   This is a typical PID control logic diagram, where the input signal is the set target, and the output signal is the error signal obtained by comparing the FEEDBACK PATH with the set input. Then, by performing proportional, integral, and differential operations on the error signal, a new input is obtained and re-entered into the system.

The traditional perspective is that the original system is an open-loop system, requiring a feedback loop to form a closed-loop system. According to this view, any system, regardless of its initial configuration, is considered open-loop. This contradicts the definition of a closed-loop system. The underlying issue, however, is that the original system is irreversible, preventing us from obtaining the appropriate adjustment signal through backpropagation based on the system's causal logic. As a result, an additional feedback loop is needed to complete the reverse process.

Nevertheless, Figure 4 provides a valuable insight: to effectively adjust the parameters within a closed system, adjustments must be made simultaneously along three different directions to achieve optimal results. This explains why, in Figure 3, even though two parameters were adjusted simultaneously, satisfactory results were not obtained. Therefore, the key to resolving the system training challenge lies in implementing a distributed PID control strategy within a distributed system.

### 4.3 Distributed PID control strategy

Traditional PID control is based on modeling the overall system, which allows for system control without the need to analyze its intricate details. Instead, it suffices to recognize the system as causal, thus bypassing the need for complex analysis. However, this approach has its limitations. When PID control is applied to the entire system, it sacrifices strong generalization capabilities. Moreover, a closed system implies that any external input is merely a projection of an external state within the system. From a mathematical standpoint, a closed system is considered complete, yet this does not imply that the system's output fully captures all relevant information. As a result, additional systems must be integrated to expand the system's state space. This concept is central to our approach of replacing digital neurons with systems in this study.

To implement PID control within a group of differential equations, we must adapt the PID method accordingly. This differs from traditional PID control because we are working within a closed system, necessitating an approach that accounts for the specific characteristics of such systems rather than simply applying open-loop system principles[7]. Given that we have three distinct sets of parameters, we assign different control strategies to each. Specifically, we apply the integral control method to $K_1$, the differential control method to $K_2$, and the proportional control method to $K_3$

For $K_1$, an integral control method is adopted

$$G_{1\_k1} = \sum_{i=1}^{kn} \int_0^T D_i(t)dt \cdot \sum_{i=1}^{kn} \int_0^T \hat{D}_i(t)dt \qquad (4.2)$$

$$G_{2\_k1} = atan(G_{1\_k1} * kt) / kt \qquad (4.3)$$

$$K_{1\_new} = K_{1\_old} \cdot exp(G_{2\_k1}) \qquad (4.4)$$

As we can see, a similar method is adopted to adjust $K_1$. The difference is that the calculation method of $G_1$ is different. In the integration strategy, we add all the signals in a single neuron together. $kn$ is the number of elements in a single neuron. This is how the integration strategy is implemented in a closed system.

For $K_2$, a differential control method is adopted:

$$G_{1\_k2} = \int_0^T D(t)dt \cdot \int_0^T \hat{D}(t)dt \cdot inputnode \qquad (4.5)$$

$$G_{2\_k2} = atan(G_{1\_k2} * kt) / kt \qquad (4.6)$$

$$K_{2\_new} = K_{2\_old} \bullet exp(-G_{2\_k2}) \qquad (4.7)$$

In 4.5, *inputnode* is a variable. If this node is an input node in the forward propagation process, then this variable is equal to 1, otherwise it is equal to 0. Because in the system, the generation of the signal is always caused by the input node, from the causal logic, the input node is in front of other nodes, which is similar to the differential control method.

For $K_3$, a proportional control method is adopted:

$$G_{1\_k3} = \int_0^T D(t)dt \bullet \int_0^T \hat{D}(t)dt \qquad (4.8)$$

$$G_{2\_k3} = atan(G_{1\_k3} * kt)/kt \qquad (4.9)$$

$$K_{3\_new} = K_{3\_old} \bullet exp(-G_{2\_k3}) \qquad (4.10)$$

The proportional control method is the easiest to understand. According to Equation 2.3, changing $K_3$ has the most direct impact on the fixed point. Therefore, we use the method of changing $K_3$ as the proportional control strategy. Although we designed three different control strategies, their efficiencies are different due to the network structure and system structure, and this design method is not the only one. More research is needed to determine their adjustment methods.

Fig. 5 Accuracy curves under different PID control strategies

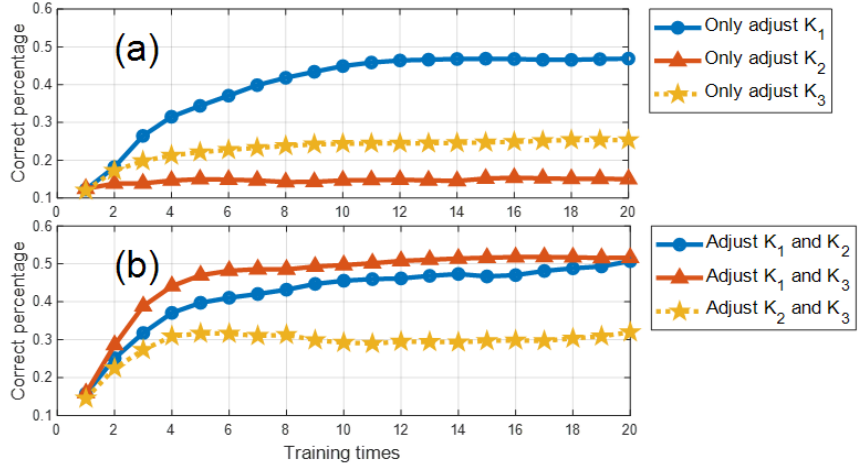

Fig. 5.a Integral strategy, differential strategy and proportional strategy are used to adjust K1, K2 and K3 respectively. It can be seen that the integral strategy has the best effect because the adjustment of the integral strategy is effective for 5 parameters at the same time, so it has a faster adjustment characteristic. The adjustment effect of the proportional strategy is slower. The differential strategy is the worst because the differential adjustment is effective for some part of the parameters.

Fig. 5.b The integral strategy, differential strategy and proportional strategy are combined for adjustment. It can be seen that compared with the original accuracy, the combined results are improved, proving that the combined parameter adjustment is effective. In actual operation, we did not choose to enable the parameter adjustment of the three strategies at the same time, mainly because the differential method has great instability. This is also the case in other industrial controls.

In Figure 5a, we use three different strategies for adjustment. It can be seen that the integral method has the best adjustment effect, because the integral method adjusts all parameters in $K_1$ at a large amplitude at the same time. The differential method has the worst adjustment effect, also because the number and amplitude of parameters adjusted by the differential method are small, and the differential method has great instability. This is also the case in other industrial controls[8]. Of course, no matter which method is used in PID, common problems in PID conditions will be encountered, such as the choice of rapidity and stability. At this time, we can refer to other methods of PID control.

In Figure 5.b, we use a combination of two strategies to adjust the parameters. It can be seen that compared with adjusting a single parameter, the accuracy and speed are significantly improved. However, when the system reaches a certain accuracy, the growth becomes slower, which is related to the network structure and parameter range limitations. We have no plans to discuss these issues in this article. In addition, theoretically, the simultaneous use of PID can achieve the best control effect, but because the differential control method used in this article is limited by the method itself, the efficiency is not high. Under current conditions, we recommend integral and proportional control. In future work, we will demonstrate more efficient differential control methods.

## 5. Summary

In this paper, we present a novel training method for the Wuxing neural network, incorporating innovations in both mathematics and biology. From a mathematical perspective, we propose the use of differential equations in place of chain derivation to address the one-to-one correspondence between parameters and outcomes. From a biological standpoint, we adopt the concept of instinctive design, limiting the adjustment process to individual neurons, which enhances the biological interpretability of the system. Building on this foundation, we further introduce a distributed PID method, applying PID control theory to individual neurons, with promising results. The use of the distributed PID method not only provides a new approach to neural network training but also offers fresh insights into the development of automatic control theory.

In this article, we expand on the method of replacing the function of a single neuron with a system as a whole. We argue that a closed system is complete in itself, and this completeness enables the system to fully respond to signal characteristics. However, this response cannot be directly observed from the external environment. Consequently, multiple systems must be interconnected, and a distributed PID control strategy must be employed to enhance the system's generalization capabilities. This approach forms the core of our perspective on neural networks.

Our research aims to establish a general neural network architecture. While the operational modes of individual neurons are diverse, this paper focuses on one particular mode for the sake of clarity. Within this framework, training methods are also varied. Here, we introduce one such method, though we have identified four other effective training strategies, which will be discussed in future work. In the next stages of our research, we plan to explore how combining different training methods can further enhance accuracy, with parameter adjustments to be addressed in subsequent studies.

## Acknowledgements

Thanks to China Scholarship Council (CSC) for their support during the pandemic, which allowed me to get through those difficult days and give me the opportunity to put my past ideas into practice, ultimately resulting in the article I am sharing with you today.

## Reference

1. Jiang, K. *A Neural Network Framework Based on Symmetric Differential Equations*. 2024. DOI: http://dx.doi.org/10.12074/202410.00055.
2. Hinton, G., *The forward-forward algorithm: Some preliminary investigations.* arXiv preprint arXiv:2212.13345, 2022.
3. Ang, K.H., G. Chong, and Y. Li, *PID control system analysis, design, and technology.* IEEE transactions on control systems technology, 2005. **13**(4): p. 559-576.

4. Hopfield, J.J., *Neural networks and physical systems with emergent collective computational abilities.* Proceedings of the national academy of sciences, 1982. **79**(8): p. 2554-2558.
5. Chua, L.O. and L. Yang, *Cellular neural networks: Theory.* IEEE Transactions on circuits and systems, 1988. **35**(10): p. 1257-1272.
6. Aihara, K., T. Takabe, and M. Toyoda, *Chaotic neural networks.* Physics letters A, 1990. **144**(6-7): p. 333-340.
7. Lombana, D.A.B. and M. Di Bernardo, *Distributed PID control for consensus of homogeneous and heterogeneous networks.* IEEE Transactions on Control of Network Systems, 2014. **2**(2): p. 154-163.
8. Åström, K.J. and T. Hägglund, *The future of PID control.* Control engineering practice, 2001. **9**(11): p. 1163-1175.